\documentclass[final]{cvpr}

\usepackage{times}
\usepackage{epsfig}
\usepackage{graphicx}
\usepackage{amsmath}
\usepackage{amssymb}
\usepackage{booktabs}
\usepackage{subfigure}
\usepackage{float}

\usepackage[pagebackref=true,breaklinks=true,colorlinks,bookmarks=false]{hyperref}

\def\cvprPaperID{3181} 
\def\confYear{CVPR 2021}

\begin{document}

\title{Domain-Specific Suppression for Adaptive Object Detection}
\author{Yu Wang\textsuperscript{1,2,3}\qquad Rui Zhang\textsuperscript{1,2}\qquad Shuo Zhang\textsuperscript{2}\qquad Miao Li\textsuperscript{1,2,3}\qquad YangYang Xia\textsuperscript{2} \\ XiShan Zhang\textsuperscript{1,2}\qquad ShaoLi Liu\textsuperscript{2}\\ 

\textsuperscript{1}SKL of Computer Architecture, Institute of Computing Technology, CAS, Beijing, China\\
\textsuperscript{2}Cambricon Technologies, China \qquad\textsuperscript{3}University of Chinese Academy of Sciences, China\\

{\tt\small \{wangyu19g,zhangrui,limiao18g,zhangxishan\}@ict.ac.cn}\\
{\tt\small \{zhangshuo,xiayangyang,liushaoli\}@cambricon.com}

}
\maketitle

\begin{abstract}
Domain adaptation methods face performance degradation in object detection, as the complexity of tasks require more about the transferability of the model. We propose a new perspective on how CNN models gain the transferability, viewing the weights of a model as a series of motion patterns. The directions of weights, and the gradients, can be divided into domain-specific and domain-invariant parts, and the goal of domain adaptation is to concentrate on the domain-invariant direction while eliminating the disturbance from domain-specific one. Current UDA object detection methods view the two directions as a whole 
while optimizing, which will cause domain-invariant direction mismatch even if the output features are perfectly aligned. In this paper, we propose the domain-specific suppression, an exemplary and generalizable constraint to the original convolution gradients in backpropagation to detach the two parts of directions and suppress the domain-specific one. We further validate our theoretical analysis and methods on several domain adaptive object detection tasks, including weather, camera configuration, and synthetic to real-world adaptation. Our experiment results show significant advance over the state-of-the-art methods in the UDA object detection field, performing a promotion of $10.2\sim12.2\%$ mAP on all these domain adaptation scenarios. 
\end{abstract}

\section{Introduction}
Deep neural networks(DNN) has achieved a tremendous breakthrough at various computer vision tasks on public dataset, including classification\cite{krizhevsky2017imagenet}, object detection\cite{girshick2014rich} and segmentation\cite{7913730}. Nevertheless, most of these researches are based on the hypothesis that the training dataset and application scenarios have identical distribution, which is apparently impossible to satisfy in practice. Unsupervised domain adaptation(UDA) provides an alternative to solve the performance degradation caused by domain distribution mismatch problem\cite{donahue2014decaf} without the need for annotations on the target domain.
\begin{figure}[t]
  \subfigure[]{ 
    \label{before} 
    \begin{minipage}[b]{0.3\linewidth} 
      \centering 
      \includegraphics[width=1\linewidth]{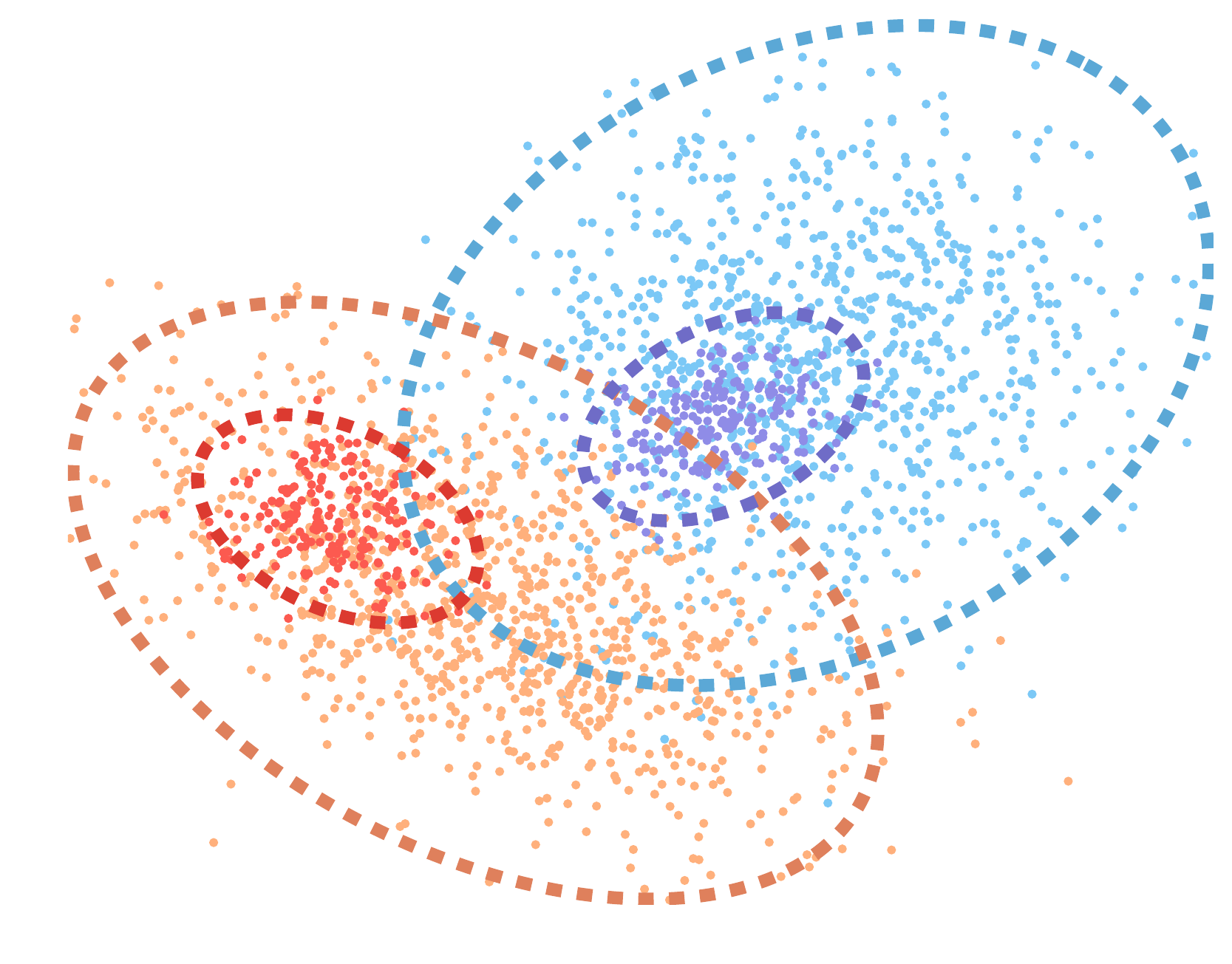} 
    \end{minipage}}%
  \subfigure[]{ 
    \label{da} 
    \begin{minipage}[b]{0.3\linewidth} 
      \centering 
      \includegraphics[width=1.1\linewidth]{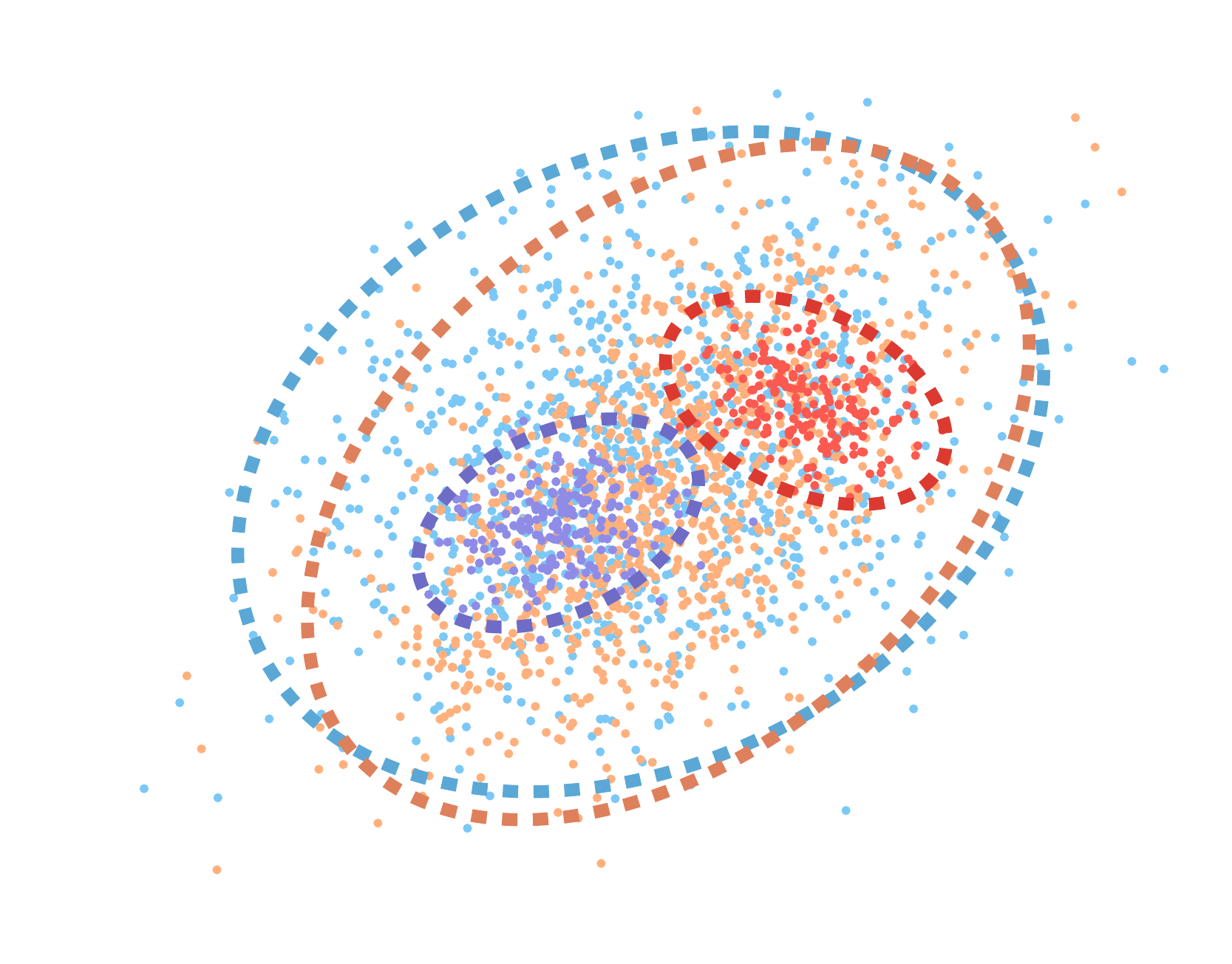} 
    \end{minipage}} 
  \subfigure[]{ 
    \label{nec} 
    \begin{minipage}[b]{0.3\linewidth} 
      \centering 
      \includegraphics[width=1.1\linewidth]{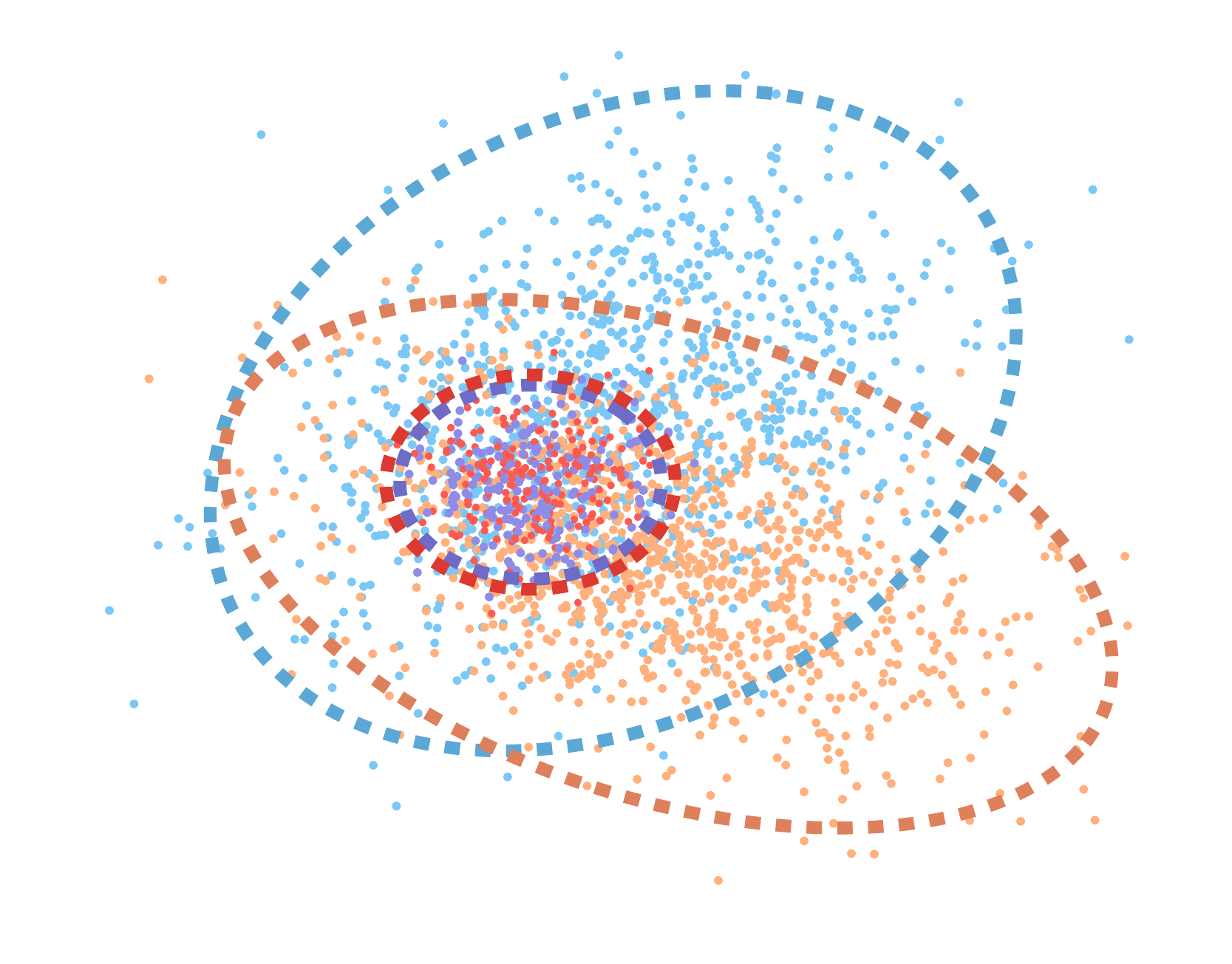} 
    \end{minipage}} 
  \caption{\textbf{Feature Distribution Sketches.} (a) represents the original feature distributions of two different domains. Features inside each domain's inner circle represent those extracted by the domain-invariant part of the model; others are from the domain-specific part of the model. (b) shows an extreme common feature alignment method, with the overall distribution aligned while the domain-invariant part mismatched. (c) represents an ideal adapted distribution, with the domain-invariant part well-aligned without the intervention of domain-specific part}

\label{distribution}
\end{figure}
We delved into the transferability of the model from the perspective on model-level explanation in the training process. We divided the weights and gradients of a model into two separate directions: domain-invariant and domain-specific. 
The former emphasizes the consistency between different domains for high-level tasks, and plays a crucial role in the transferability of a model. In contrast, the latter is the main obstacle to the transferability of a model, as it represents the unique features within a certain domain which have no concern with the tasks. In this way, an ideal domain adaptation method is expected to promote transferability by learning the domain-invariant weights while eliminating the domain-specific one.

Current research in UDA strives to align the distribution of features extracted by the model with auxiliary objective function measuring the discrepancy between them. With the common goal, early research tries to measure and minimize the distance of features from different domains in a well-designed feature space, such as maximum mean discrepancy(MMD)\cite{borgwardt2006integrating}, and Kullback-Leibler divergence(KL)\cite{zhuang2015supervised}. With the introduction of GAN\cite{goodfellow2014generative}, methods that measure the discrepancy between domains by one or more discriminator distinguishing the origin of features spring up. Constraint to the feature-level explanation, such trail of study considers the domain-specific and the domain-invariant parts as a whole and optimizes them together. However, all the similar distribution can guarantee is that the combinations of the domain-specific and domain-invariant part of the model for different domains are equivalent. Considering the disturbance of domain-specific part, the domain-invariant part of different domains may still be inconsistent even if the similarity condition is satisfied, as shown in Figure \ref{distribution}. The optima of domain adaptation is the alignment of domain-invariant part and the eradication of domain-specific part, as is illustrated in Figure \ref{distribution}. This is exactly the reason why current methods seem to be inefficient or even ineffective when the high-level task becomes more complicated, such as object detection in this paper.

With the purpose of domain-invariant alignment, we propose a novel domain-specific suppression (DSS) method. We roughly estimate the domain-specific part of the gradients with its projection on the direction of weights, and impose restrict to the gradients in the corresponding direction. Such estimation relies on the fact that the overall proportion of domain-specific direction in weights is generally higher than that in gradients since the gradients are dominated by domain-specific deviation initially, and there is an updating lag between the gradients and weights. The gradient will gradually converge to the domain-invariant optima with the constrain on domain-specific direction since both the domain-specific direction and the domain-invariant direction can lead to their corresponding local optima for the final task. Furthermore, we provide a special case of domain-specific suppression by normalizing the weight with its 2-Norm. Such simplification can significantly reduce the implementing consumption, making the domain-specific suppression a plug-and-play block to any architecture. With domain-specific suppression, we remove one key barrier to domain adaption tasks.

We evaluate our method on the Faster RCNN framework, ResNet-50 backbone on various datasets: Cityscapes, Foggy Cityscapes, KITTI, and SIM10K, involving weather variance, camera configuration changes and synthesize to real-world adaptations. We further implement additional experiments with the model pre-trained on the COCO2017 to illustrate the necessity of improving the model's discriminability by pre-training on a large dataset in UDA detection task. With DSS, we have outperformed state-of-the-art methods on all domain adaptation object detection tasks with various datasets. This achievement means that our methods have almost bridged the gap between two domains with a simple distribution mismatch.
\section{Related Work}
\textbf{Object Detection}
Object Detection is one of the fundamental tasks in computer vision. Current object detection methods based on CNN can be roughly divided into two types: one-stage object detectors\cite{liu2016ssd}\cite{lin2017focal} and two-stage object detectors\cite{xiang2017subcategory}\cite{cai2016unified}. With R-CNN\cite{2014Rich} pioneers the way of extracting region proposals with selective search, the two-stage R-CNN series detectors become one of the most popular object detectors. R. Girshick\cite{Girshick_2015_ICCV} proposed Fast RCNN, accelerate training process by sharing the convolution output features with RoI pooling. Followed by Fast R-CNN, Faster R-CNN\cite{ren2015faster} came out as the first end-to-end detector, proposing a region proposal network to generate region proposal with almost no-cost.
R. Joseph et al.\cite{redmon2016you} proposed the first one-stage detector YOLO series\cite{redmon2017yolo9000}\cite{redmon2018yolov3}\cite{bochkovskiy2020yolov4}. Completely abandoned the pattern of verification and regression, YOLO predicts bounding boxes directly from images by regression alone.

\textbf{Domain Adaptation}
Domain adaptation has been a hot spot and widely researched on classification tasks for a long time, and there are excessive achievements\cite{hu2015deep}\cite{tzeng2015simultaneous}\cite{lu2017unsupervised}\cite{zellinger2017central}. In the early stage, research prefers to design a latent space to measure the discrepancy of features extracted from source and target domains. The most widely used measurements are maximum mean discrepancy(MMD)\cite{tzeng2014deep}\cite{long2015learning}\cite{long2017deep}\cite{long2016unsupervised} and Wasserstein distance\cite{shen2017wasserstein}. Since the introduction of GAN, a lot of UDA methods based on adversarial learning spring up\cite{bousmalis2017unsupervised}\cite{deng2018image}\cite{tzeng2017adversarial}\cite{wang2020progressive}\cite{yu2019transfer}\cite{chen2019progressive}\cite{long2018conditional}\cite{DBLP:journals/tip/JiaoYX21}. Different from the former one, adversarial domain adaptation measures the discrepancy of features by one or several discriminators dynamically in the training process. With the guide of feature distance measurement, deep neural networks are supposed to learn a domain invariant feature space.

\textbf{UDA in Object Detection}
Researches on domain adaptation in object detection are still in the early stage\cite{li2016weakly}\cite{Inoue_2018_CVPR}\cite{khodabandeh2019robust}\cite{saito2019strong}, the first of which goes back to\cite{xu2014domain}, applying an adaptive SVM to the deformable part-based model(DPB). \cite{raj2015subspace} firstly attempted to introduce two-stage detectors in domain adaptation by aligning the feature subspace of R-CNN. Inducting the achievement of domain adaptation in classification, \cite{chen2018domain} proposed a pioneering framework of domain adaptive Faster RCNN, embedding adversarial block into Faster RCNN detection, aligning both image-level and instance-level features. Followed by this work, lots of adversarial-based domain adaptation frameworks in object detection arise.\cite{chen2020harmonizing}\cite{zhu2019adapting}\cite{hsu2020progressive}\cite{xu2020exploring}\cite{kim2019diversify}

\section{Preliminaries}
We briefly revisit the forward and backward process of a fully-connected neural network and its properties. Denoting the input of any layer as $z$, and the output as $y$, the forward process is a combination of function composition and matrix multiplication:
\begin{equation}
    y=f(Wz),
\end{equation}
where $W$ is a $m\times n$ real matrix. $f(\cdot)$ refers to activation function like ReLU or Sigmoid. A singular value decomposition of matrix $W$ is given by:
\begin{equation}
    W=U\Sigma V,
\end{equation}
where $U,V$ are real matrices with $m\times m$ and $n\times n$ dimension respectively. $\Sigma$ is an $m\times n$ rectangular diagonal matrix with non-negative real numbers on the diagonal, known as the singular values of $W$. The weight matrix $W$ can be interpreted as a linear transformation that rotates the input vector $z^l$ from one orthogonal basis $U$ to another orthogonal basis $V$ with different scaling in each direction (and a projection if $n\neq m$). 

In this way, the forward process of each layer can be viewed as a motion pattern applied to the input features, pointing to a corresponding direction of the weight matrix. Furthermore the back-propagation in each layer can be interpreted as an updating of the transforming direction, adjusting the motion pattern pointing to the direction of the optimum feature space for a certain task.

As convolution layer is a special case of fully connected layer with weight reusing and multiple channels, we can easily extend this to convolutional neural networks.
\subsection{Consistency, Specificity and Transferability}
\begin{figure}[t]
\begin{center}
\includegraphics[width=0.5\textwidth]{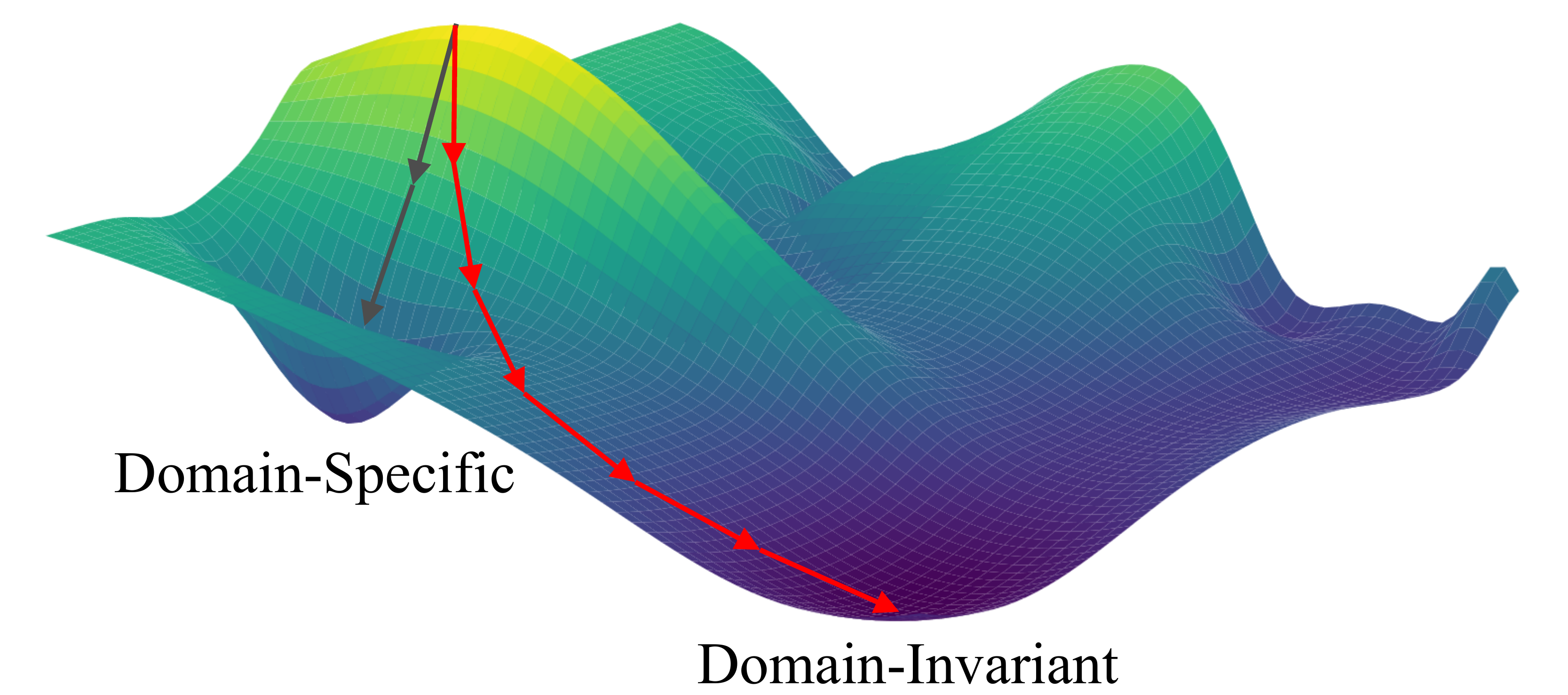}
\end{center}
   \caption{We illustrate two kinds of optimization processes with input distribution mismatch. The black path represents an original method that will converge to the domain-specific optima guided with the whole gradients directly. The direction of solid red line is the optimizing route of the optimization process with domain-specific suppression, which is supposed to find the domain-invariant optima.}
\label{loss}
\end{figure}

Generally speaking, the features captured by shallow layers of a deep neural network are low-level appearance, containing more information about edges, textures, etc. The features from deep layers, in contrast, containing more about high-level semantic information. Intuitively, those peculiar properties reflect a logical or causative connection with the transferability of neural networks. 

Based on such observation, we investigate such phenomena from a different perspective: we set our sight on the motion patterns of the model. As a neural network can be viewed as a series of motion patterns transforming the input to a task-friendly feature space, the entire process can be roughly divided into two phases: decomposing the input into essential features and generating semantic features for the final tasks.  

Obviously, the model's transferability, namely the ability of a model to perform well across different domains, is primarily dictated by the first phas as the model will perform well on the target domain with only the annotations in the source domain if the model can extract consistent essential features. However, once the essential features extracting pattern of the model contains too much domain bias, nothing the second phase can do to perform well on the target domain. On the other hand, whether the destination of the first phase is specific or consistent, the second phase can always find its optimum to the final task for the domain with annotations. For example, a domain within which all the cars are red may cause misunderstanding to the model that cars should be in red, but such misunderstanding will not result in any performance degradation within the exact domain compared with a model without such color bias. 

We divide the gradients backpropagating in the training process into two separate directions: domain-invariant and domain-specific, emphasizing the consistency and specificity of domains, respectively. As the DNN has been revealed a particular preference in capturing dataset bias in the internal representation \cite{tommasi2017deeper}, the motion patterns in the first phase are sensitive to the domain-specific direction. As shown in Figure \ref{loss}, the gradients to the domain-specific optima are far more sharp than to the domain-invariant one. That is to say while updating, the speed to the domain-specific direction is much higher than that of the domain-invariant direction without any extra constraint. 

According to the analysis above, an ideal domain adaptation method is expected to eliminate the domain-specific direction in the first phase and find the direction from domain-invariant feature space to a task-friendly feature space in the second phase. 

\subsection{Limitation in Current UDA object detection framework}
Current research on domain adaptations in object detection has a typical pattern of constructing feature spaces with the output feature maps of the detectors (whether from backbone or ROI-head) to measure the discrepancy of domains and then diminishing or interpolating it. Based on the analysis above, it is easy to find that rather than optimizing the domain-invariant direction (eliminating the domain-specific direction in the meantime), such a method is regarding the domain-specific direction and the domain-invariant one as a whole. As shown in Figure~\ref{loss}, the motion pattern of the model will be heavily influenced by the domain-specific direction, entrapped in the domain-specific optima as the black path. While the model can still perform well on the source domain, the performance in the target domain is bound to be unsatisfying with the lack of annotations.

\section{Method}

Based on the theoretical analysis above, we propose an exemplary and generalizable solution to eliminate the domain-specific direction at its root.

\subsection{Domain-Specific Suppression}

\begin{figure}[t]
\begin{center}
\includegraphics[width=0.4\textwidth]{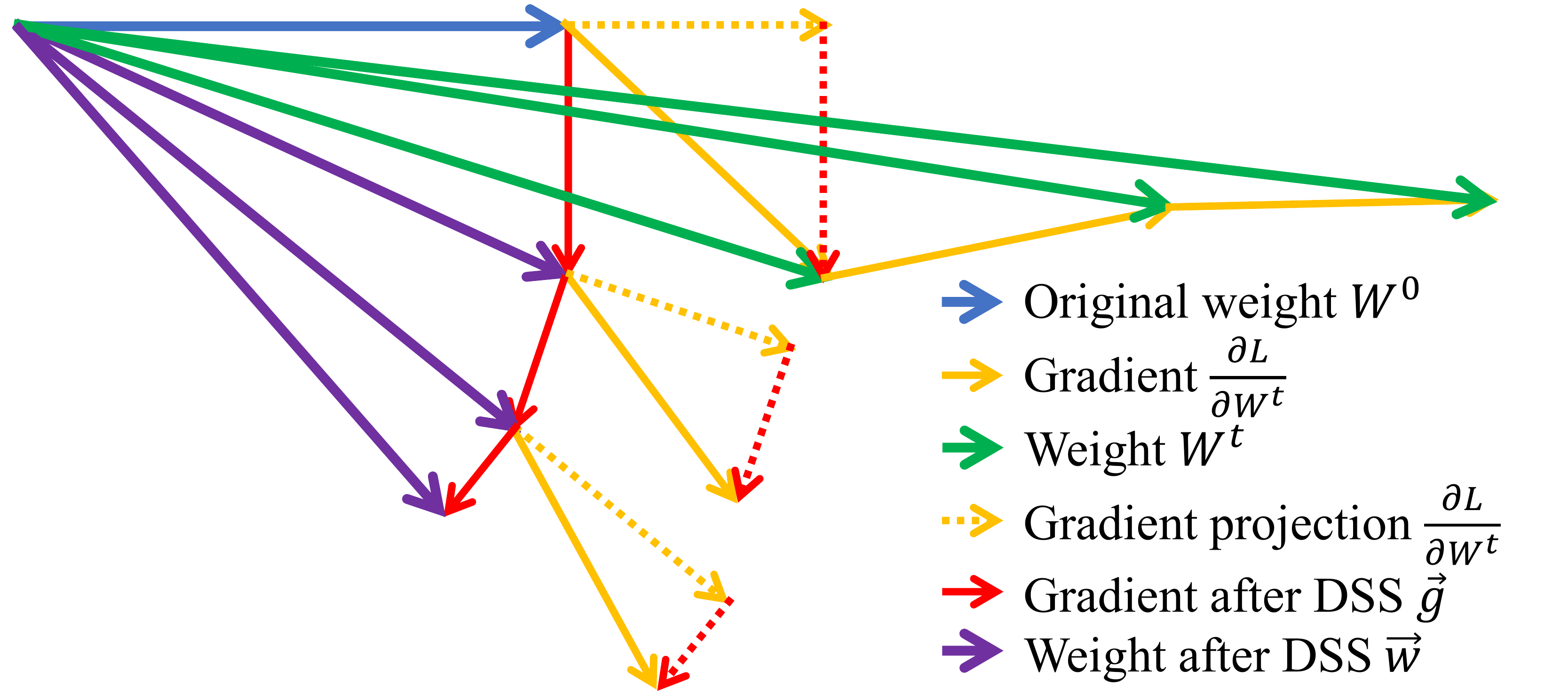}
\end{center}
   \caption{\textbf{Weights Updating Process.} We present three steps of weights updating with and without DSS.}
\label{route}
\end{figure}

The Deep neural network adjusts the series of motion patterns it applies to the input features by backpropagation, which can be written as the following:
\begin{equation}
    W^{t+1} = W^{t} - \eta*\frac{\partial L}{\partial W^{t}},
\end{equation}
where $W^{t}$ means the weight matrix in the $t-th$ iteration. Such a backward process treats domain-invariant and domain-specific directions of the gradients equivalent, optimizing the weights with both of them, which will lead to the aforementioned domain-specific optima.

Motivated by the analysis above, our domain-specific suppression is supposed to separate the domain-specific direction in gradients and eliminate its influence on the training process. First, we estimate the domain-specific direction with the direction of the weights of the model, and then eliminate it by subtracting the projection of gradients on the direction of weights, adding constraint terms to the gradients during the whole training process as following: 
\begin{equation}
\label{equ4}
\begin{split}
     W^{t+1}_i =&W^{t}_i - \eta*(\frac{\partial L}{\partial W^{t}}\\
     &-\lambda<\frac{\partial L}{\partial W^{t}},\frac{W^{t}}{\sqrt{\Vert W^{t} \Vert_2^2}}>\cdot \frac{W^{t}}{\sqrt{\Vert W^{t} \Vert_2^2}})\\
     =& W^{t}_i - \eta*(\frac{\partial L}{\partial W^{t}}-\lambda<\frac{\partial L}{\partial W^{t}},W^{t}>\cdot \frac{W^{t}}{\Vert W^{t} \Vert_2^2}).
\end{split}
\end{equation}
Figure \ref{route} illustrate the updating process with equation~\eqref{equ4}, where $\frac{W^t}{\sqrt{\Vert W^t\Vert_2^2}}$ is the direction of weight $W^t$, $<\frac{\partial L}{\partial W^{t}},\frac{W^{t}}{\sqrt{\Vert W^{t} \Vert_2^2}}>$ is the norm of the projection of gradient $\frac{\partial L}{\partial W^{t}}$ on the direction of weight $W^t$, and then\newline $<\frac{\partial L}{\partial W^{t}},\frac{W^{t}}{\sqrt{\Vert W^{t} \Vert_2^2}}>\cdot \frac{W^{t}}{\sqrt{\Vert W^{t} \Vert_2^2}}$ represents the projection of gradient $\frac{\partial L}{\partial W^{t}}$ on the direction of weight $W^t$.

We estimate and eliminate the domain-specific direction by the constraint in equation~\eqref{equ4} based on the following analysis. As the model trained by UDA framework shows a better performance in the source domain than the target one, the original motion patterns are dominated by the source domain-specific direction naturally. In this way, the weights of model at the beginning can exactly indicate the direction of the domain-specific direction we are hoping to alleviate. During the training process, the direction of the whole model gradually collaborates, and the domain-invariant direction will prevail over the specific one, consequently. Considering the lagging in updating, the ratio of domain-specific direction in weights is always higher than that in gradients. In this way, what is left after the subtraction remains to be gradients with a lower ratio of domain-specific direction.
That is to say, the higher the relation between the direction of gradients and weights, the lower the domain-invariant information that gradient provides. In this case, the motion patterns will be updated in a direction with the domain-specific direction being suppressed.

Domain-specific suppression can significantly improve the performance of a domain adaptation method, especially when the model is well-pretrained on the source domain. The reason is that the constraint will curb the learning of the final task to some degree, but a pre-training process on the source domain will make up for the limitation.
\begin{figure*}
\begin{center}
\includegraphics[width=0.75\textwidth]{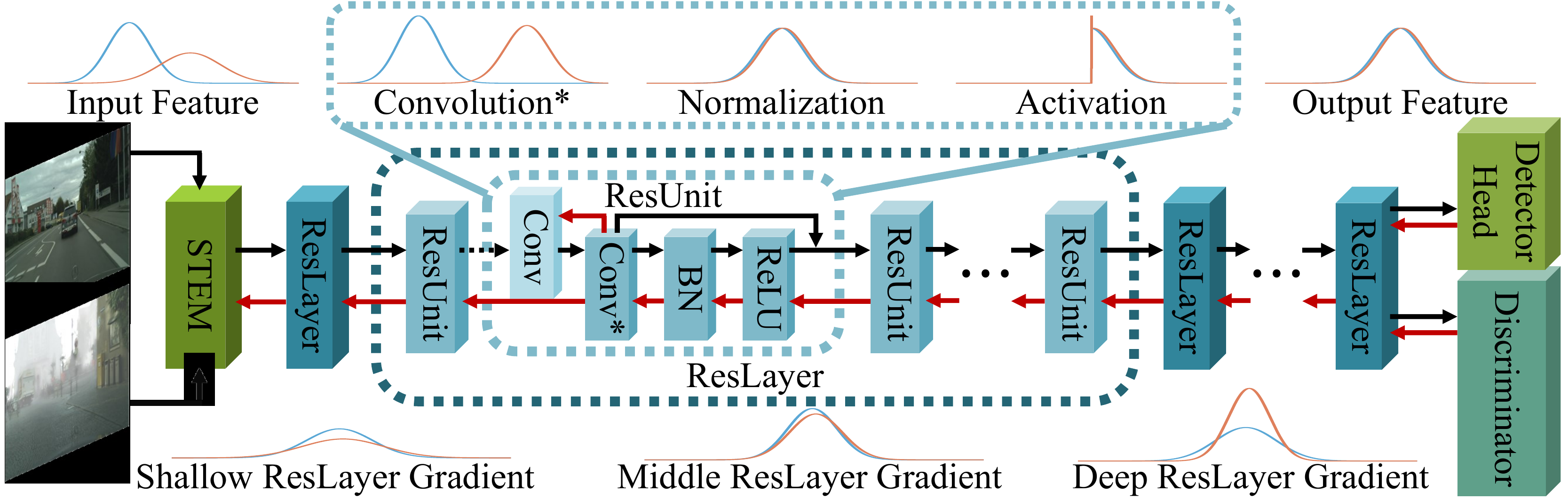}
\end{center}
   \caption{\textbf{Domain Adaptation Framework with Domain-Specific Suppression Block.} We illustrate the transforming trend of input features after different kinds of layers above the framework. The histograms behind are gradient distributions of gradients from layers of different depths}
\label{framework}
\end{figure*}

\subsection{A special Domain-Specific Suppression case: Normalization with Frobenius Norm}
We provide a more implementing-friendly domain-specific suppression by normalizing the weight of each convolution layers by Frobenius Norm. This can be easily inserted into any architectures. We will prove its equivalence with domain-specific suppression.

Assuming a network A with weight matrix $\Omega=\{\omega_i\}$. In the training process, A uses $\widetilde{\Omega}=\frac{\Omega}{\Vert \Omega \Vert}$ in forward propagation, then we have the backpropagation as following:
\begin{equation}
\begin{split}
\frac{\partial L}{\partial \omega_i} &= \frac{\partial L}{\partial \widetilde{\omega_i}}\frac{\partial \widetilde{\omega_i}}{\partial \omega_i} + \sum_{j\neq i}\frac{\partial L}{\partial \widetilde{\omega_j}}\frac{\partial \widetilde{\omega_j}}{\partial \omega_i}\\
&=\frac{\partial L}{\partial \widetilde{\omega_i}}(\frac{1}{\sqrt{\Vert \Omega\Vert_2^2}}-\frac{\omega_i^2}{(\sqrt{\Vert \Omega \Vert_2^2})^3}) \\
&\quad + \sum_{j\neq i}\frac{\partial L}{\partial \widetilde{\omega_j}}(-\frac{\omega_i\omega_j}{(\sqrt{\Vert \Omega \Vert_2^2})^3})\\
&=\frac{1}{\sqrt{\Vert \Omega\Vert_2^2}}\frac{\partial L}{\partial \widetilde{\omega_i}}-\frac{1}{\sqrt{\Vert \Omega \Vert_2^2}}\widetilde{\omega_i}\sum_{j}\frac{\partial L}{\partial \widetilde{\omega_j}}\widetilde{\omega_j},
\end{split}
\end{equation}
\begin{equation}
    \frac{\partial L}{\partial \Omega}=\frac{1}{\sqrt{\Vert \Omega \Vert_2^2}}\frac{\partial L}{\partial \widetilde{\Omega}} - \frac{\widetilde{\Omega}}{\sqrt{\Vert \Omega \Vert_2^2}}<\frac{\partial L}{\partial \widetilde{\Omega}}, \widetilde{\Omega}>.
\label{equ6}
\end{equation}
As the actual weights updated in backpropagation is $\Omega$, we have the backward process as following:
\begin{align}
\Omega^{t+1}&=\Omega^{t} - \eta * \frac{\partial L}{\partial \Omega^{t}},
\end{align}
We have the valid assumption in training process that $\sqrt{\Vert \Omega^{t} \Vert_2^2}\approx \sqrt{\Vert \Omega^{t+1} \Vert_2^2}$, then we have
\begin{equation}
\begin{split}
    \widetilde{\Omega}^{t+1}&=\frac{\Omega^{t+1}}{\sqrt{\Vert \Omega^{t+1}\Vert_2^2}}\\
    &\approx\frac{\Omega^{t}-\eta * \frac{\partial L}{\partial \Omega^{t}}}{\sqrt{\Vert \Omega^{t}\Vert_2^2}}\\
    &=\frac{\Omega^{t}}{\sqrt{\Vert \Omega^{t} \Vert_2^2}}-\eta*\frac{\partial L}{\partial \Omega^{t}}*\frac{1}{\sqrt{\Vert \Omega^{t} \Vert_2^2}}.
\end{split}
\end{equation}

Combined With the equation~\eqref{equ6}, the updating process of $\widetilde{\Omega}$ can be written as:
\begin{equation}
\begin{split}
    \widetilde{\Omega}^{t+1}&=\frac{\Omega^{t}}{\sqrt{\Vert \Omega^{t} \Vert_2^2}}-\eta*\frac{1}{\Vert \Omega^{t} \Vert_2^2}(\frac{\partial L}{\partial \widetilde{\Omega}^t}-\widetilde{\Omega}^t<\frac{\partial L}{\partial \widetilde{\Omega}^t},\widetilde{\Omega}^t>)\\
    &=\widetilde{\Omega}^{t} - \hat{\eta}*\frac{1}{\Vert \Omega^{t} \Vert_2^2}(\frac{\partial L}{\partial \widetilde{\Omega}^t}-\lambda\widetilde{\Omega}^t<\frac{\partial L}{\partial \widetilde{\Omega}^t},\widetilde{\Omega}^t>).
\end{split}
\end{equation}
Then we have the backpropagation in domain-specific suppression format as:
\begin{align}
    &\Vert \widetilde{\Omega}^t\Vert = 1,\\
    &\lambda=1,\\
    &\hat{\eta}=\frac{\eta}{\Vert \Omega^{t} \Vert_2^2}.
\end{align}
Then 2-Norm can be cast as a special case of equation~\eqref{equ4} with  $\lambda = 1$ and an adaptive  $\hat{\eta}$. 
\subsection{UDA Object Detection Framework}
The general pipeline of our UDA object detection framework is shown in figure \ref{framework}. The whole framework consists of a backbone network followed by two parallel parts: a detector head and a domain discriminator. Our domain-specific suppression modifies the backbone network with its convolution layers. The Conv layer represents the weights $\Omega$ above, and the Conv* layer represents the $\widetilde{\Omega}$. In the training process, the actual weights used in the forward process is calculated from Conv layer, while in the backward process, the weights in Conv layer will be updated with $\frac{\partial L}{\partial \widetilde{\Omega}}\frac{\partial \widetilde{\Omega}}{\partial \Omega}$. 

\begin{table*}[ht]
\begin{center}
\begin{tabular}{c|cccccccc|c}
\toprule[2pt]
Methods & Person & Rider & Car & Truck & Bus & Train & Motorbike & Bicycle & mAP \\
\hline
Source Only&17.8&23.6&27.1&11.9&23.8&9.1&14.4& 22.8& 18.8\\
\hline
DA-Faster\cite{chen2018domain}&25.0&31.0&40.5&22.1&35.3&20.2&20.1&27.1&27.6\\
SCDA\cite{zhu2019adapting}&33.5&38&48.5&26.5&39&23.3&28&33.6&33.8\\
DivMatch\cite{kim2019diversify}&35.1&42.1&49.1&30.0&45.2&26.9&26.8&36.0&36.4\\
Progressive DA\cite{hsu2020progressive}&36.0&45.5&54.4&24.3&44.1&25.8&29.1&35.9&36.9\\
SWDA\cite{saito2019strong}&32.9&43.8&49.2&27.2&45.1&36.4&30.3&34.6&37.4\\
HTCN\cite{chen2020harmonizing}&33.2&47.5&47.9&31.6&47.4&40.9&32.3&37.1&39.8\\
\hline
DSS(Source Only)&46.2&50.5&53.2&25.9&43.4&21.2&33.1&45.0&39.8\\				
DSS(UDA Framework)&42.9&51.2&53.6&33.6&49.2&18.9&36.2&41.8&40.9\\
\hline
Source Only*&42.3&49.9&45.0&23.2&35.4&16.5&32.2&41.5&35.8\\
DSS(Source Only)*&50.9&57.6&61.1&35.4&50.9&36.6&38.4&51.1&47.8\\
DSS(UDA Framework)*&\textbf{50.0}&\textbf{58.6}&\textbf{66.5}&\textbf{36.1}&\textbf{57.1}&\textbf{50.0}&\textbf{44.5}&\textbf{53.0}&\textbf{52.0}\\
\bottomrule[2pt]
\end{tabular}
\end{center}
\caption{Results of object detection adapting from Cityscapes to Foggy Cityscapes(Weather Adaptation) Methods with * represents that the model has been pre-trained on COCO before finetune on source domain.}

\label{cityscapes}
\end{table*}
During the forward process of domain-specific suppression, the input features with different distributions will be transformed into Gaussian distributions with similar standard deviations but different means with the convolution layer(Conv*). The deviations of means will be further eliminated by the normalization layers(Batch Normalization, for example). The final output features for the following detector head and domain discriminator will have no appreciable difference. While in the backward process, the gradients in the shallow layer will be amplified as those layers need more adjustment for a domain-invariant direction. The gradients in the deep layers, however, will be suppressed, as they are supposed to concentrate more on semantic information, which is consistent between different domains.

Our method can be inserted into any other UDA object detection framework without extra consumption.
\section{Experiments}
In this section, we validate our method on typical kinds of domain discrepancy:1. Weather discrepancy 2. Camera setting discrepancy 3. Synthesis to real-world discrepancy. 

We provide an additional MS COCO pre-trained baseline to validate that domain-invariant direction is consistent between different domains, and DSS can learn it effectively.

We also provided further theoretical analysis of our method with insight experiments, including gradient quantitative study and convergence comparison in speed and accuracy between domains.

\subsection{Dataset}
\textbf{Cityscapes}
Cityscapes dataset is a large-scale city street scene dataset collected from different cities. It contains 2975 training images and 500 validation images with 8 classes ground truth labels.\newline
\textbf{Foggy Cityscapes}
Foggy cityscapes dataset is a synthetic dataset derive from the Cityscapes dataset. The synthetic foggy transmittance is generated automatically, inheriting the semantic annotation of original images. Each image in Cityscapes will be added with fog in three density levels, so this dataset contains 8925 training images and 1500 validation images.\newline
\textbf{SIM10K}
SIM10K are synthetic datasets generated by the grand theft auto(GTAV) engine. It contains 10000 images with 58071 bounding box annotations, with only car in the category. We divided it randomly into 8000 images for training and 2000 images for validation.\newline
\textbf{KITTI}
KITTI is one of the most significant datasets in the self-driving field. Images are collected from rural, urban and highway areas in a driving car. KITTI contains 7481 images with annotations. We set full of the dataset as the training set.
\subsection{Experiment Details}
We utilize ResNet-50 as the backbone, and the parameters of the backbone are initialized from the model pre-trained on ImageNet and fine-tuned on corresponding source domains. For each iteration, one batch of source domain input and one batch of target domain input will be fed into the model simultaneously, while the target domain dataset will only contribute to the domain discriminator loss. We set the default batch size of each domain as 2 per GPU. We evaluate mean average precisions with a threshold of 0.5 to compare with other methods.
\subsection{Results}

\subsubsection{Weather Adaptation}
\textbf{Settings}
In this section, we utilize Cityscapes as the source domain and Foggy Cityscapes as the target domain. In the source domain, all images in Cityscapes are used with annotations, while in target domain only images will participate in the training process. Adaptation results are evaluated on the validation set of Foggy Cityscapes with all eight categories.\newline \textbf{Results}
The results are presented in Table\ref{cityscapes}. As shown in the results that our method achieves an equivalent performance with the state-of-the-art methods when training only on the source domain. This exactly validates that our method can eliminate the influence of domain-specific direction effectively. When inserted into a UDA framework, our method can gain an extra 1$\%$ mAP improvement, which means that information from the target domain can further refine the domain-invariant direction.

In additional COCO pre-trained set, the results of DSS show an incredible improvement of 8$\%$ mAP comparing with state-of-the-art methods with source domain alone, and achieve 52$\%$ mAP when inserted into a basic UDA framework. This considerable improvement proves our analysis that the side-effect of DSS that will suppress the learning of object detection tasks can be reduced significantly. The results also illustrate that pre-train alone can not distinguish the domain-invariant direction accurately, and its help in transferability is limited. The promotion with DSS validates that domain-invariant direction is consistent between different domains and DSS helps to learn it better with COCO.
\begin{figure}[h]
  \subfigure[]{ 
    \label{grada} 
    \begin{minipage}[b]{0.49\linewidth} 
      \centering 
      \includegraphics[width=1\linewidth]{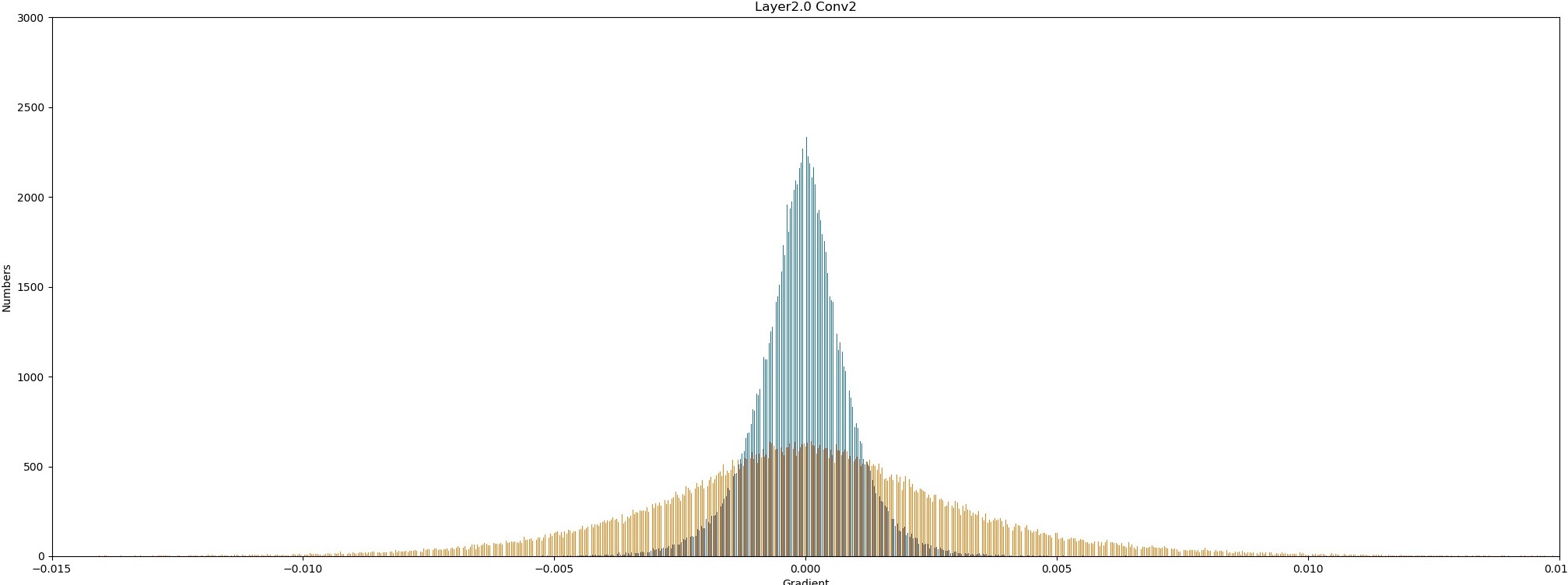} 
    \end{minipage}}%
  \subfigure[]{ 
    \label{gradb} 
    \begin{minipage}[b]{0.48\linewidth} 
      \centering 
      \includegraphics[width=1\linewidth]{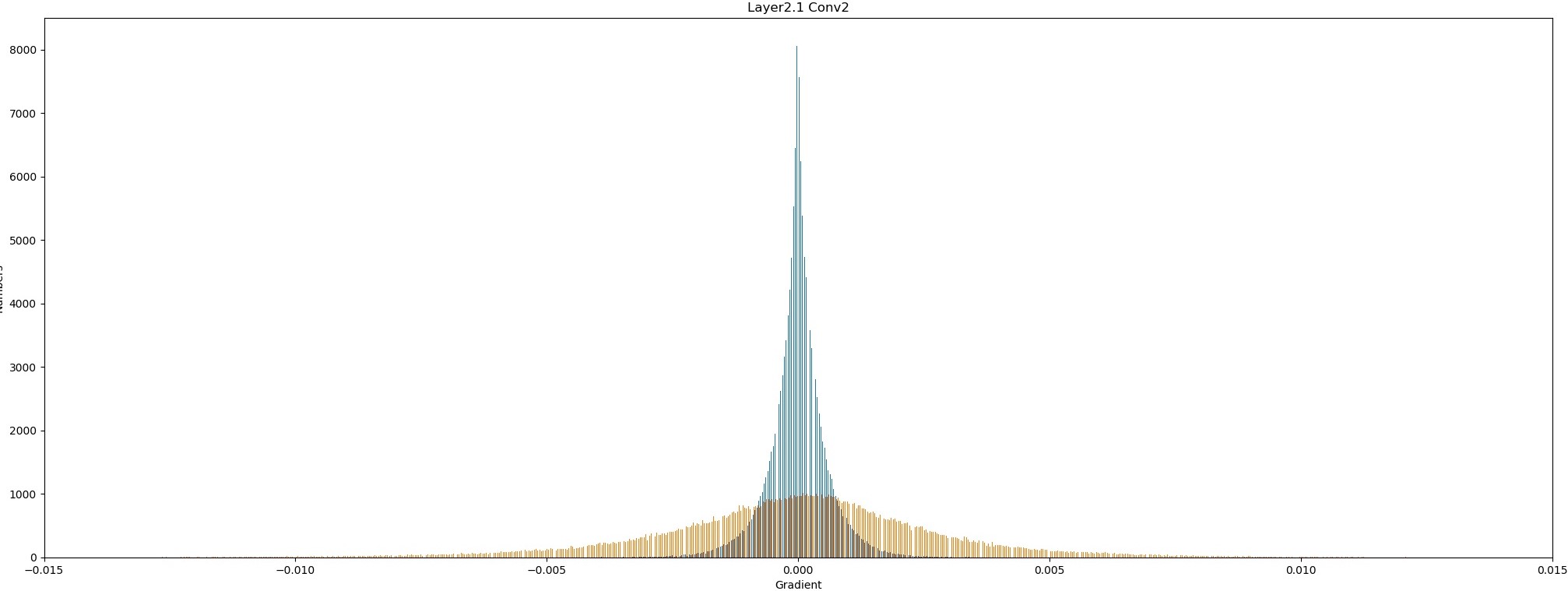} 
    \end{minipage}} 
  \subfigure[]{ 
    \label{gradc} 
    \begin{minipage}[b]{0.48\linewidth} 
      \centering 
      \includegraphics[width=1\linewidth]{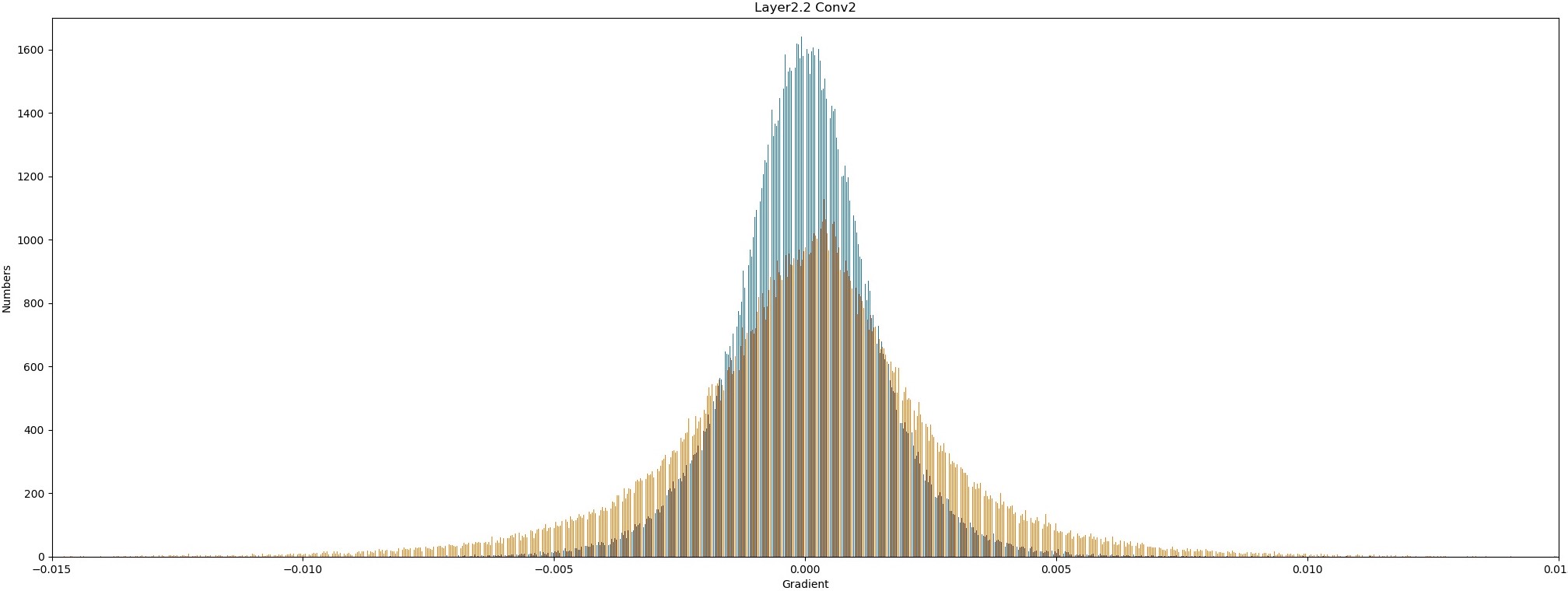} 
    \end{minipage}} 
  \subfigure[]{ 
    \label{gradd} 
    \begin{minipage}[b]{0.48\linewidth} 
      \centering 
      \includegraphics[width=1\linewidth]{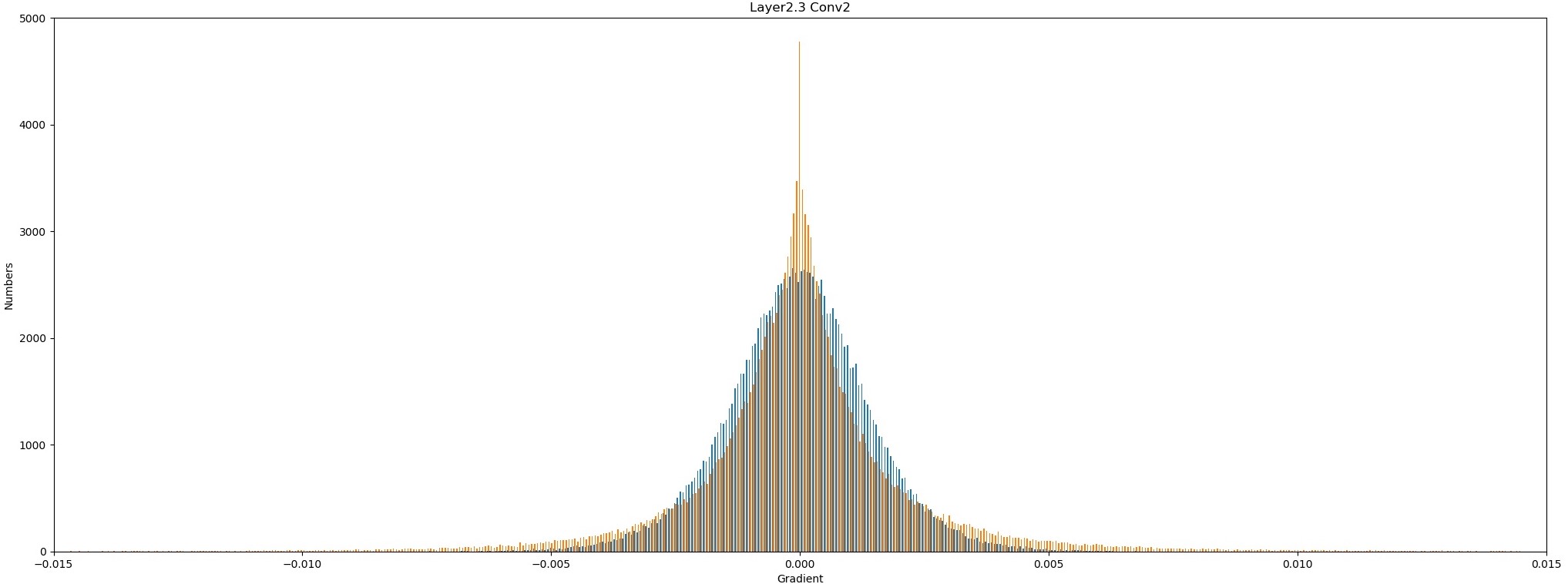} 
    \end{minipage}} 
  \subfigure[]{ 
    \label{grade} 
    \begin{minipage}[b]{0.48\linewidth} 
      \centering 
      \includegraphics[width=1\linewidth]{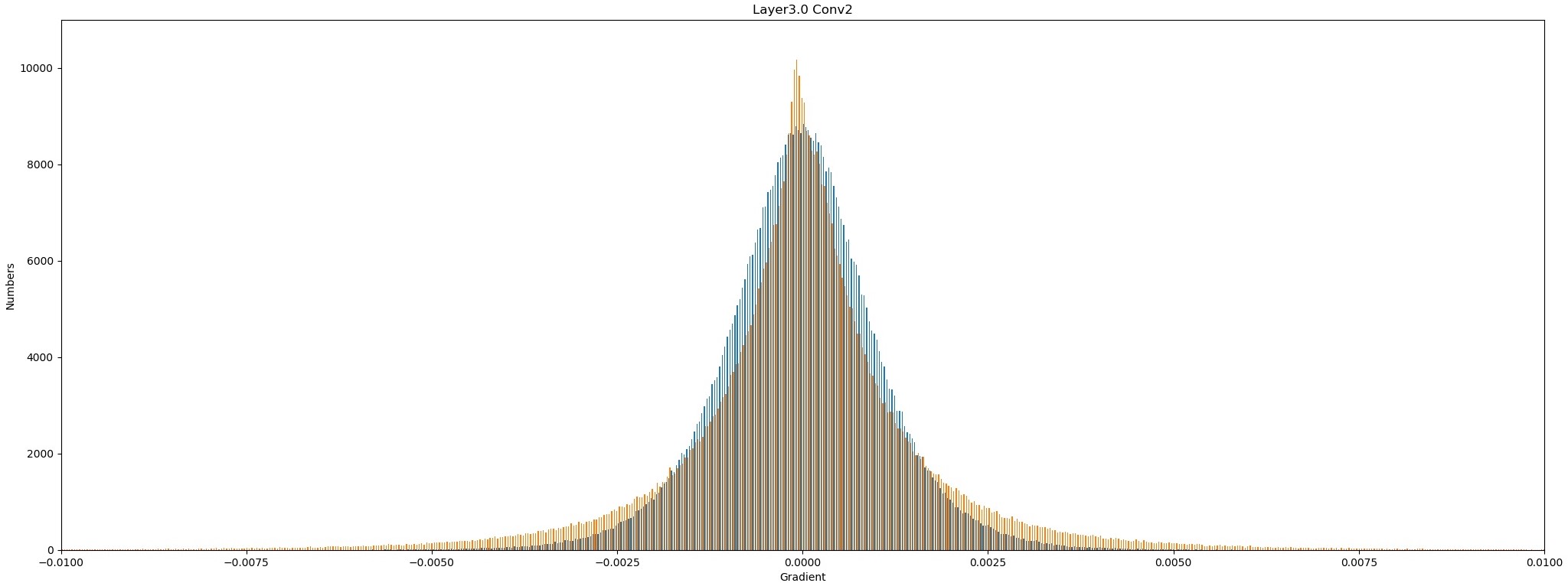} 
    \end{minipage}} 
  \subfigure[]{ 
    \label{gradf} 
    \begin{minipage}[b]{0.48\linewidth} 
      \centering 
      \includegraphics[width=1\linewidth]{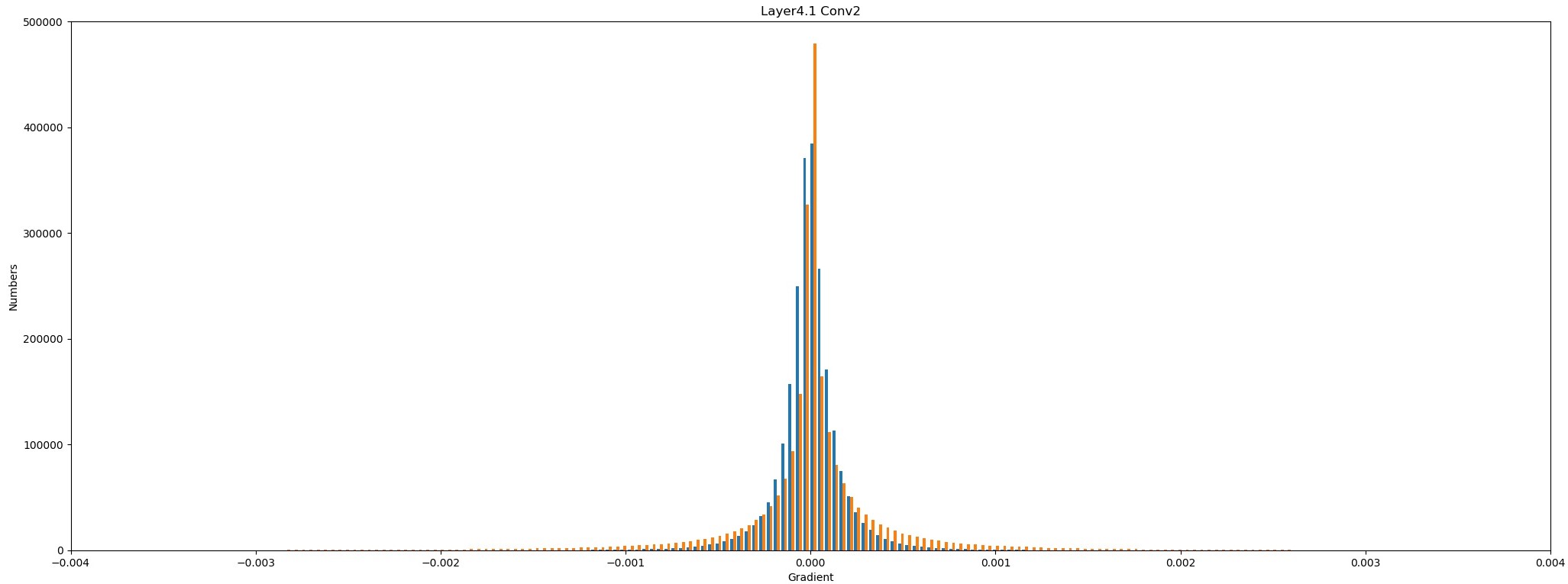} 
    \end{minipage}} 
  \caption{\textbf{Gradient Histogram Comparison} Fig \ref{grada}-\ref{gradf}present the histograms of gradients in backbone(ResNet-50) during training process with(blue lines) and without(yellow lines) domain-specific suppression. Gradients are extracted from the first $3\times 3$ convolution layers of ResLayer 2.0, 2.1, 2.2, 2.3, 3.0, 4.1 respectively.} 
  \label{grad} 
\end{figure}
\subsubsection{Camera Configuration Adaptation}
\textbf{Settings}
In this section, we utilize KITTI as the source domain and Cityscapes as the target domain to evaluate the performance of our method in camera configuration mismatch scenarios. The whole KITTI dataset and the training set of Cityscapes will be utilized in the training process. Results are evaluated on the only common category between two domains: car.\newline
\textbf{Results}
Comparison with current state-of-the-art methods is presented in Table\ref{table:kitti}. While our method imporves 26.5$\%$ mAP in KITTI to Cityscapes task when pre-trained with COCO, the performance of DSS alone seems to be ordinary. Analyzing this phenomenon from the perspective on the discrepancy between this pair of domains, we can find that the performance's bottleneck is not the transferability of the model but the discriminability. Statistics shows that the average instances per image are 4.3 in KITTI but 18 in Cityscapes. The model will face more complicated semantic confusion, such as overlapping and sheltering, which requires more of the discriminability of the model. The considerable improvement brought by pre-trained routine also validates that once we make up for the shortcomings of discriminability of the model, our domain-specific suppression can improve the performance of a model to a higher level.
\begin{table}[t]
\begin{center}
\begin{tabular}{c|c}
\toprule[2pt]
Methods  & Car mAP \\
\hline
Source Only&34.6\\
\hline
DA-Faster\cite{chen2018domain} & 41.9\\
HCTN\cite{chen2020harmonizing} & 42.5\\
SCDA \cite{zhu2019adapting}& 43.0\\
\hline
DSS(Source Only)&41.6\\				
DSS(UDA Framework)&42.7\\
\hline
Source Only*&39.8\\
DSS(Source Only)*&42.6\\
DSS(UDA Framework)*&\textbf{59.2}\\
\bottomrule[2pt]
\end{tabular}
\end{center}
\caption{Results of object detection adapting from KITTI to Cityscapes(Camera configuration Adaptation). Methods with * represents that the model has been pretrained on COCO before fine-tuned on source domain.}
\label{table:kitti}
\end{table}

\subsubsection{Synthesis to Real World Adaptation}

\textbf{Settings}
Such adaptation is meaningful as generating synthesis data can reduce the cost of sampling and labelling remarkably. We utilize SIM10K as the source domain and Cityscapes as the target domain. Results are evaluated with the only common category: car between two domains.\newline
\textbf{Results}
The final results are shown in Table\ref{table:SIM10K}. Our DSS provides performance close to state-of-the-art methods without pre-trained routine while achieves 1.9$\%$ mAP improvement over current methods and 10.7$\%$ promotion when inserted into UDA framework. This proves that our method is robust concerning the pattern and texture distribution mismatches.
\begin{table}[t]
\begin{center}
\begin{tabular}{c|c}
\toprule[2pt]
Methods  & Car mAP \\
\hline
Source Only&34.7\\
\hline
DA-Faster\cite{chen2018domain} & 38.5\\
SCDA\cite{zhu2019adapting}& 42.5\\
Progressie DA\cite{hsu2020progressive} & 43.9\\
\hline
DSS(Source Only)&42.0\\				
DSS(UDA Framework)&44.5\\
\hline
Source Only*&39.3\\
DSS(Source Only)*&49.8\\
DSS(UDA Framework)*&\textbf{58.6}\\
\bottomrule[2pt]
\end{tabular}
\end{center}
\caption{Results of object detection adapting from SIM10K to Cityscapes(synthesis to real-world adaptation). Methods with * represent that the model has been pretrained on COCO before fine-tuned on source domain.}
\label{table:SIM10K}
\end{table}
\begin{figure*}[h]
  \subfigure[]{ 
    \label{s1} 
    \begin{minipage}[s1]{0.33\linewidth} 
      \centering 
      \includegraphics[width=1\linewidth]{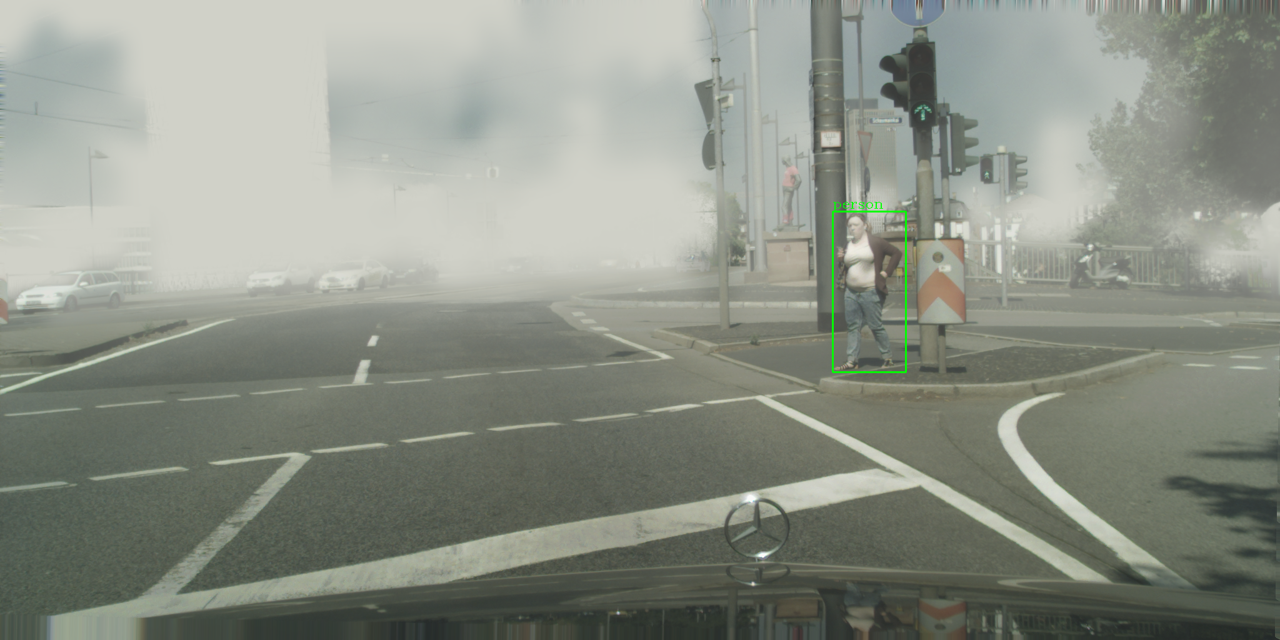} 
    \end{minipage}}%
  \subfigure[]{ 
    \label{b1} 
    \begin{minipage}[b1]{0.33\linewidth} 
      \centering 
      \includegraphics[width=1\linewidth]{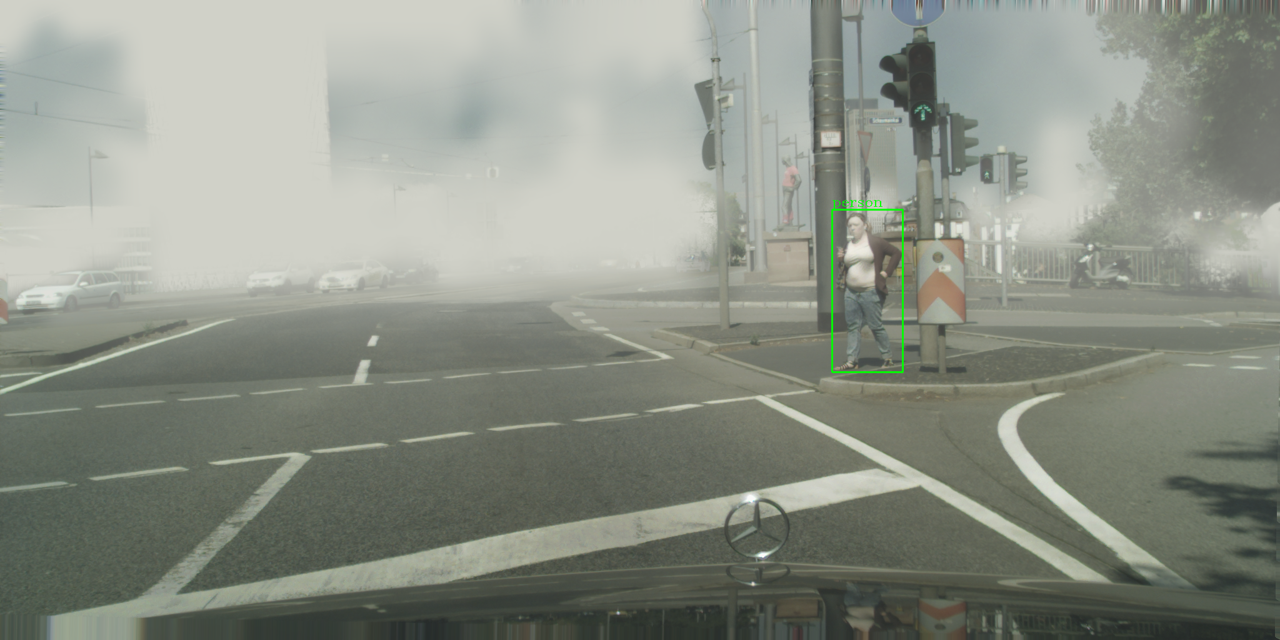} 
    \end{minipage}} 
  \subfigure[]{ 
    \label{w1} 
    \begin{minipage}[w1]{0.33\linewidth} 
      \centering 
      \includegraphics[width=1\linewidth]{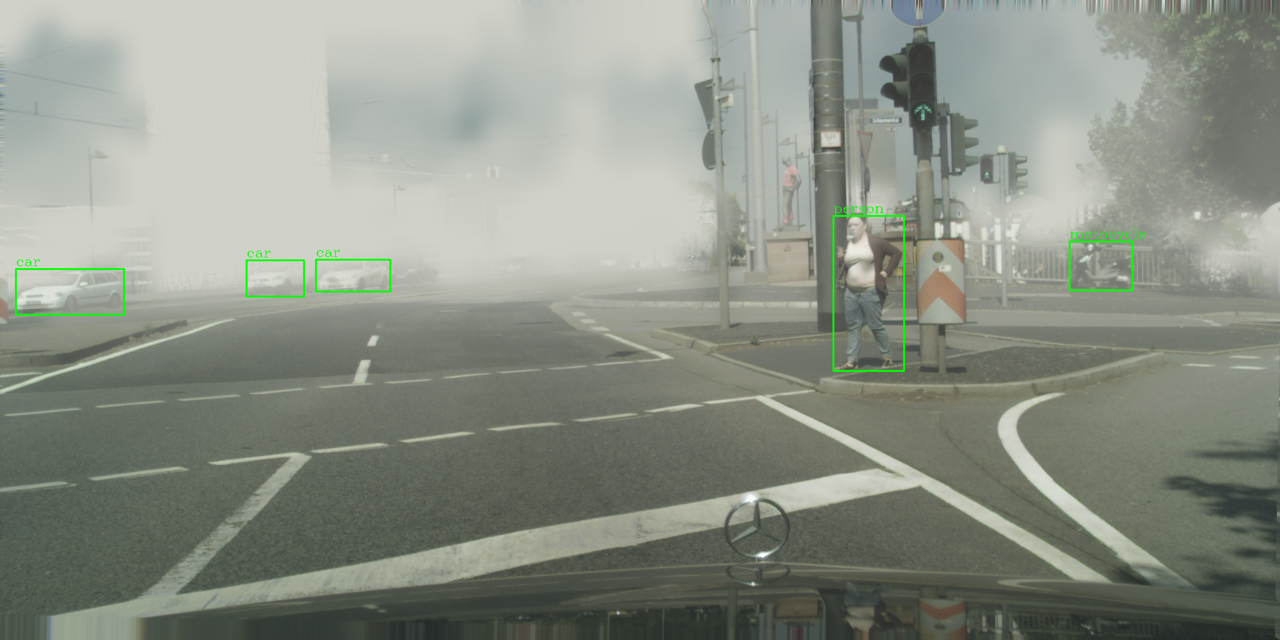} 
    \end{minipage}} 
  \subfigure[]{ 
    \label{s2} 
    \begin{minipage}[s2]{0.33\linewidth} 
      \centering 
      \includegraphics[width=1\linewidth]{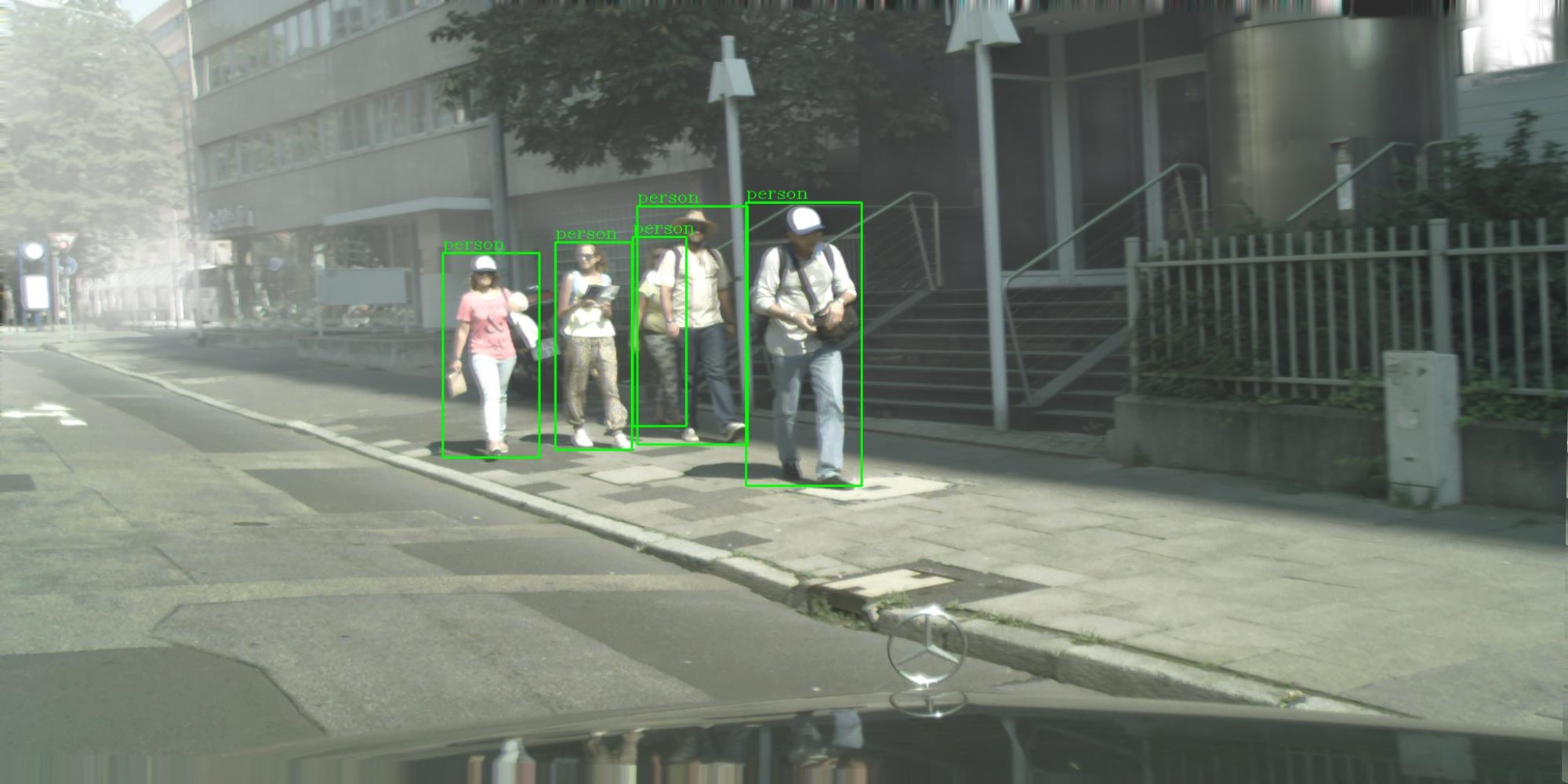} 
    \end{minipage}} 
  \subfigure[]{ 
    \label{b2} 
    \begin{minipage}[b2]{0.33\linewidth} 
      \centering 
      \includegraphics[width=1\linewidth]{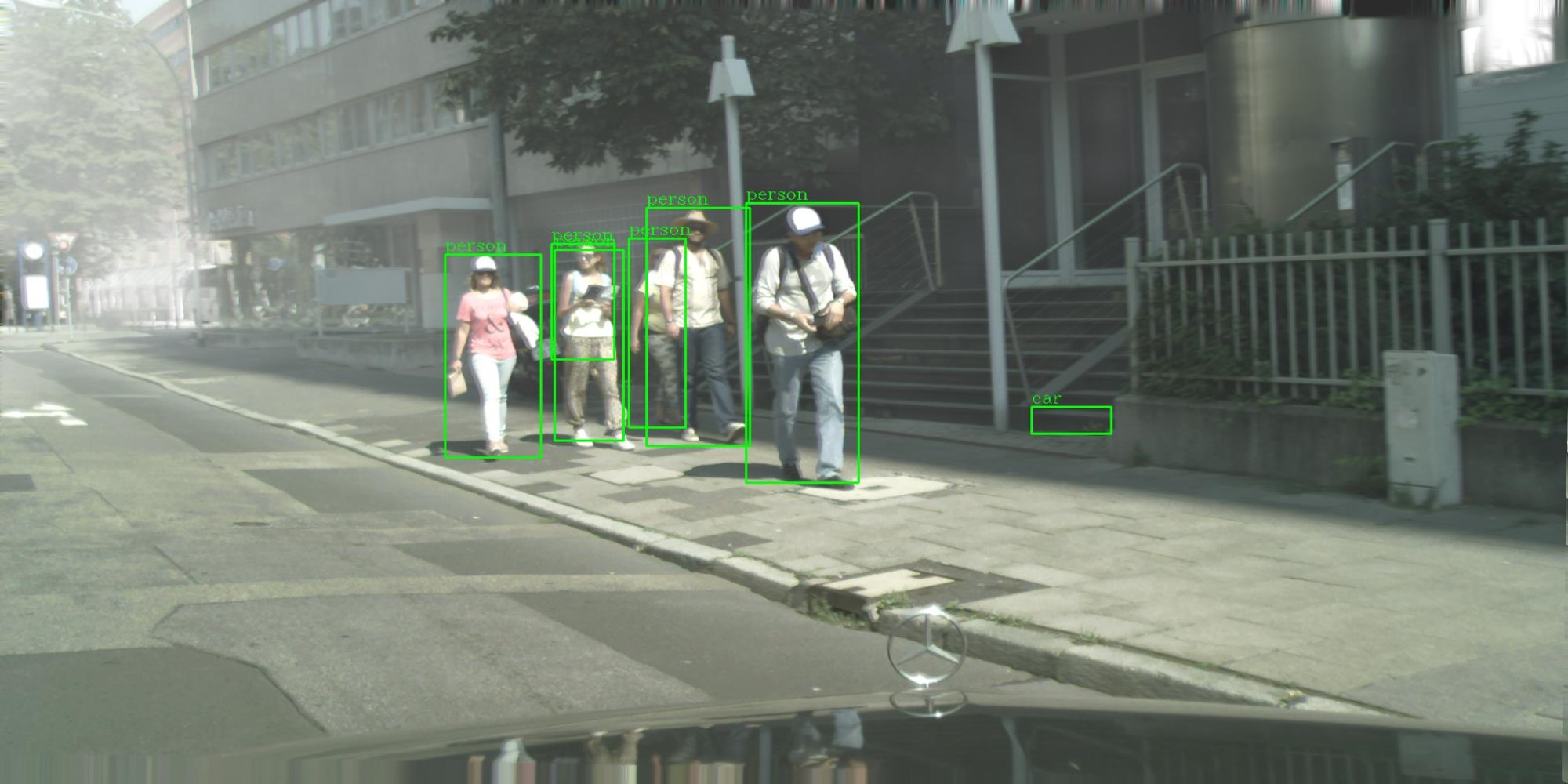} 
    \end{minipage}} 
  \subfigure[]{ 
    \label{w2} 
    \begin{minipage}[w2]{0.33\linewidth} 
      \centering 
      \includegraphics[width=1\linewidth]{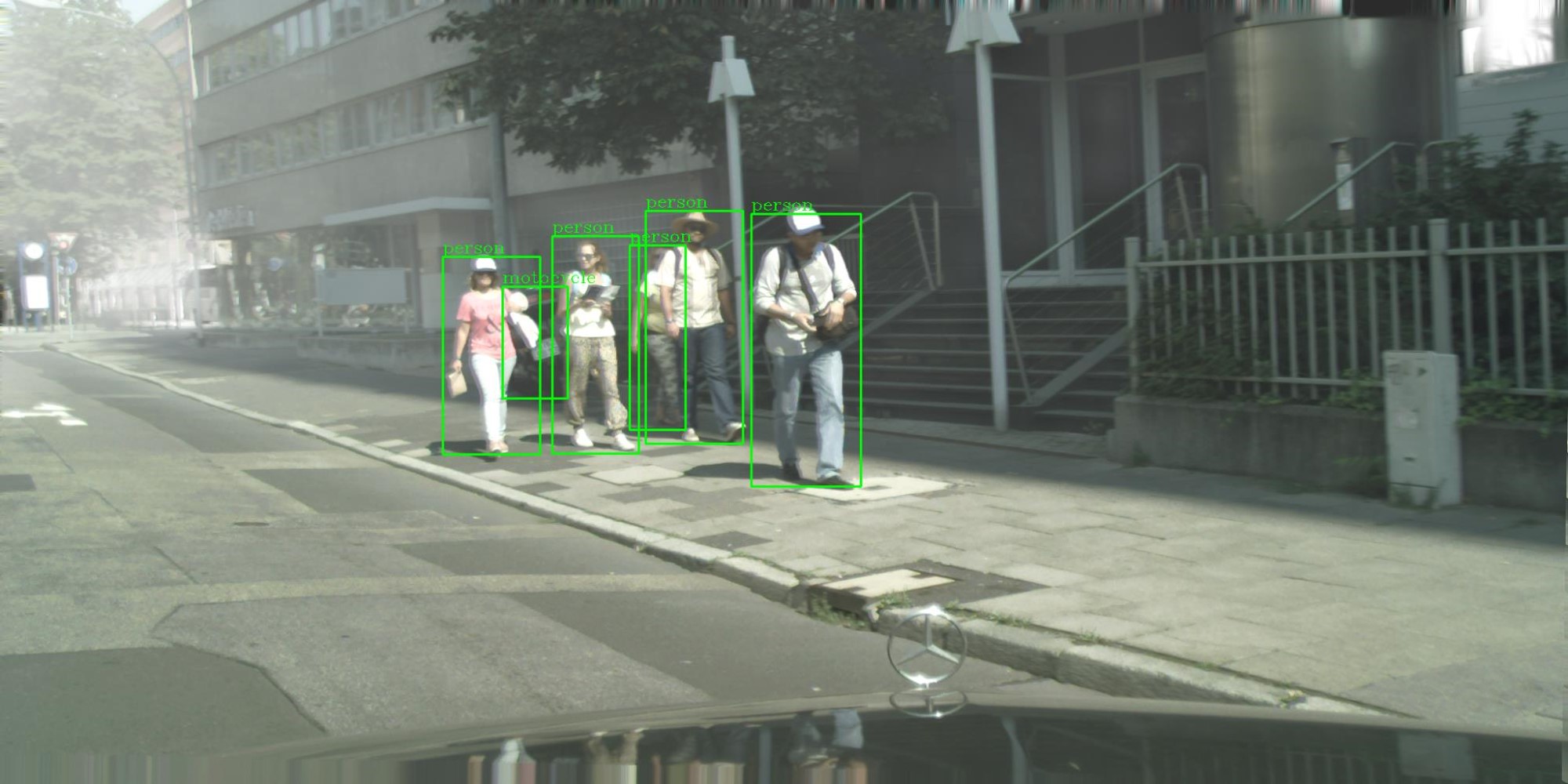} 
    \end{minipage}} 
  \caption{\textbf{Object Detection Display} Fig \ref{s1}-\ref{w1} and \ref{s2}-\ref{w2} present two set of detection results on different inputs. Results from left to rights each line are detected by source-only Faster R-CNN, UDA framework and UDA framework with DSS respectively.} 
  \label{img} 
\end{figure*}
\subsection{Gradient Quantification Study}
To fully understand the true influence of the gradient distillation constraint on the weight updating process, we compare the gradient distribution and the trends before-and-after scenarios in Figure \ref{grad}. Fig\ref{grada} and Fig\ref{gradb} show considerable increases in the scale of gradients in shallow layers, as the peaks of the histograms around zeros are significantly reduced, while the standard deviation increases remarkably. However, the effects of domain-specific suppression in  Fig\ref{gradd}~Fig\ref{gradf} are directly opposite. Gradients with domain-specific suppression in these layers are more concentrated around zeros, with higher peaks.

Obviously, after applying domain-specific suppression, the gradients of shallow layers have been strongly amplified, while the gradients of deep layers suppressed. Such phenomena exactly validate our theoretical analysis. The former methods try to constrain the combination of domain-invariant and domain-specific direction together, while our method explicitly decompose the two parts.

The comparison shows that motion patterns in shallow layers of the network are crucial to the transferability of model, and the original weights and gradients all deflect from the domain-invariant direction, trapped in the local optima for the source domain. With the suppression strategy, the updating process is forced to change direction and find a way to optimize domain-invariant feature space. Consequently, the gradient in shallow layers updates rapidly with the domain-specific suppression, collaborating its direction to the domain-invariant feature space.

While the adjustment in deep layers shows that motion patterns in deep layers concentrate more on extracting and analyzing semantic information, which can be generalized to different domain-specific feature spaces, having limited influence on the transferability of the model.

\subsection{Visualization}
\textbf{Detection Results}
We display two typical sets of detection results in Figure \ref{img}, including source-only, UDA framework and UDA framework with DSS. The first line of images illustrates the performance degradation of the DNN detector when the source and target domain have distribution mismatches caused by dense fog. While current UDA framework can mitigate the problem, our DSS can handle such a situation perfectly. In Fig \ref{b2} we illustrate a classical domain-invariant mismatch issue. With the lack of ability to domain-invariant direction accurately, the transferability of the model even decreases after feature alignment. Our DSS can solve this problem fundamentally.

\section{Conclusion}
We have presented a new perspective to understand the transferability of DNN in model-level explanation. We view the model as a series of motion patterns and divide the direction of weights and gradients into domain-invariant and domain-specific parts. The former determines the transferability of a model while the latter is an obstacle in domain adaptation. We propose domain-specific suppression to optimizing the domain-invariant parts by estimating and eliminating the domain-specific direction in gradients. Our method outperforms state-of-the-art methods for a large margin. As for future work, the relationship of discriminability and transferability of a model needs further investigation. The complicated scenarios semantic distribution mismatch may requires optimaztion both of them in combination.

\section*{Acknowledge}This work is partially supported by the National Key Research and Development Program of China(under Grant 2020AAA0103802, 2018AAA0103300), the NSF of China(under Grants 61925208, 61732007, 61732002, 61906179, U19B2019, U20A20227),Beijing Natural Science Foundation (JQ18013), Strategic Priority Research Program of Chinese Academy of Science (XDB32050200), Youth Innovation Promotion Association CAS, Beijing Academy of Artificial Intelligence (BAAI) and Beijing Nova Program of Science and Technology (Z191100001119093), Youth Innovation Promotion Association CAS and Xplore Prize.
{\small
\bibliographystyle{ieee_fullname}
\bibliography{cvpr}
}
\end{document}